# Analysis of Zero Day Attack Detection Using MLP and XAI

Ashim Dahal[1], Prabin Bajgai[1] and Nick Rahimi[1]

[1] University of Southern Mississippi, Hattiesburg MS 39406, USA
`ashim.dahal@usm.edu, prabin.bajgai@usm.edu, nick.rahimi@usm.edu`

**Abstract.** Any exploit taking advantage of zero-day is called a zero-day attack. Previous research and social media trends show a massive demand for research in zero-day attack detection. This paper analyzes Machine Learning (ML) and Deep Learning (DL) based approaches to create Intrusion Detection Systems (IDS) and scrutinizing them using Explainable AI (XAI) by training an explainer based on randomly sampled data from the testing set. The focus is on using the KDD99 dataset, which has the most research done among all the datasets for detecting zero-day attacks. The paper aims to synthesize the dataset to have fewer classes for multi-class classification, test ML and DL approaches on pattern recognition, establish the robustness and dependability of the model, and establish the interpretability and scalability of the model. We evaluated the performance of four multilayer perceptron (MLP) trained on the KDD99 dataset, including baseline ML models, weighted ML models, truncated ML models, and weighted truncated ML models. Our results demonstrate that the truncated ML model achieves the highest accuracy (99.62%), precision, and recall, while weighted truncated ML model shows lower accuracy (97.26%) but better class representation (less bias) among all the classes with improved unweighted recall score. We also used Shapely Additive exPlanations (SHAP) to train explainer for our truncated models to check for feature importance among the two weighted and unweighted models.

## 1 Introduction

A zero-day vulnerability is a flaw in a computer system or network previously unknown to its developers or anybody who could mitigate it. Any exploit taking advantage of zero-day is called a zero-day attack. Defense against any zero-day attack via patching is virtually impossible since anti-virus products are not capable of detecting the attacks through signature-based scanning with a reliable and robust accuracy (a mean value of 17% accurate detection for zero-day attacks vs 54% for theoretically known attacks) [1, 6]. This failure of traditional signature-based scanning from anti-virus calls for a different approach to detecting zero-day attacks: an Intrusion Detection System (IDS), the most common and effective way of detecting zero-day attacks [6].

Machine learning can be used to develop and scale the traditional approaches to zero-day attack detection, including heuristics, sandbox analysis, and anomaly detection. Deep learning takes this a step further by adding layers of functions to better distinguish between data from different classes. These techniques are among the most promising and scalable methods for detecting zero-day attacks [2].

Existing ML-based intrusion detection systems (IDS) often face limitations, such as high variance, imbalanced class distribution, and a trade-off between accuracy and recall. This paper proposes a novel approach to improve these limitations by leveraging supervised machine learning algorithms to develop an IDS with high accuracy, low variance, and robust performance across all classes.

This paper will train the model using four MLPs [3]. MLPs are well-suited for deep learning because they can model complex relationships, train efficiently using fully connected layers, and benefit from graphics processing unit (GPU) parallelization for classification tasks.

The remainder of this paper is structured as follows: S*ection II* describes past works and details on research on zero-day, *Section III* discusses the methodology, ML and DL algorithms and models, data preprocessing techniques, and the reasoning behind those decisions, *Section IV* presents the results and their implications in the given approach and finally, in *Section V*, the importance of the paper and its concluding observations are made.

## 2 Literature Review

Summarizing over twenty-four years of research put into IDS, powered by ML or not, with KDD99 would be beyond the scope of this paper. This paper will solely focus on overcoming the inefficiencies or discrepancies of previous research on zero-day attack detection using machine learning and deep learning techniques. Y. Guo et al. [4] provide a robust and comprehensive foundation for a survey of ML-based zero-day attack detection approaches. The review covers a wide range of topics, including the diverse types of machine learning-based zero-day attack detection methods, the challenges of machine learning-based zero-day attack detection, and the evaluation of machine learning-based zero-day attack detection systems. The research has influenced the field by identifying the key challenges and future directions for this research area. Still, it does not evaluate the performance of any specific methods. The research needs to address the issues of deploying ML-based zero-day attack detection systems in real-world situations, such as the necessity for training data and efficient and interpretable models.

I. Hairab et al. [5] evaluated the performance of convolutional neural network (CNN) models regularized by L1 and L2 regularization for detecting zero-day attacks in the ToN-IoT [6] dataset. However, this study only evaluates the proposed CNN on two types of attacks: backdoor attacks and scanning attacks. The ToN-IoT dataset contains various other attack types, such as denial-of-service attacks, heartbeat attacks, and port scanning attacks. The paper provides a promising approach for detecting zero-day attacks using CNNs. However, more research is needed to evaluate the performance of the proposed CNN model in a broader range of attack types and to address the challenges of deploying the model in real-world settings.

After closer inspection of [7,8], it can be reasonably inferred that maintaining high accuracy in a dataset with heavy class imbalance like the KDD99 [9] cannot classify any proposed model as a good approach. S. Mukunth et al. [10] proposed a support vector machines (SVM) approach for the KDD99 dataset with a high accuracy of 99.3%. However, the paper's authors do not address the class imbalance in the dataset. Similarly, S. Kumar et al. [11] achieved an accuracy of 99.5% on the KDD99 dataset using a genetic algorithm to optimize the features and parameters of an SVM classifier. However, this paper also does not address the class imbalance in the KDD99 dataset. In a dataset like the KDD99 dataset with a heavy class imbalance, classifying every data to a single class, Smurf, with 2.9 million samples, would lead to an accuracy of 60%. Therefore, after reviewing the mentioned papers, this paper aims to develop an IDS that can produce a highly accurate, interpretable, and scalable model with a strong emphasis on reducing class imbalance in KDD99, ultimately leading to a robust ML-powered IDS.

Frameworks like SHAP have helped and provided updated and transparent information to consumers regularity bodies about the reasoning behind security decisions in modern research methodology [13]. Such robustness of results provided by SHAP would help our research to solidify the reasoning behind model's predictions on the same instances.

## 3 Methodology

This section contains the proposed methodology to fulfill the research objectives described in Section I (Introduction). KDD99 [9] was chosen as the dataset for the experimentation of the ML-powered IDS. Various data pre-processing techniques and solutions were applied to the dataset.

### 3.1 Dataset

The original KDD99 dataset can be found in Kaggle, a reliable resource for obtaining many historically essential datasets. KDD99 has 4,898,431 samples with 41 features that distinguish each sample into one of 23 given classes. The dataset has a heavy class imbalance with the Smurf class, which has 2.8 million samples, while the Spy class only has two. Smurf has more samples than all other 22 classes combined; it is 60% of the entire dataset. This possesses the dataset with the problem where a lot of heavy lifting must be done to synthesize unique techniques to overcome this complication, which is discussed in Section II.

**Table 1.** Number of samples in each category after grouping.

| Attack category | Number of attacks |
|---|---|
| DoS | 3883370 |
| Normal | 972781 |
| Probe | 41102 |
| Unauthorized Access | 1178 |

### 3.2 Data Preprocessing

To reduce the class imbalance for pattern recognition and supervised machine learning algorithms, the 23 classes were further grouped into four main parent categories: probe attacks, distributed denial of service (DDoS), unauthorized access, and normal. This categorization of classes would then lead to class distribution as shown in Table 1. This organization helps reduce the huge discrepancy between the classes.

**Probe:** This category encompasses network probing and reconnaissance activities involving scanning, testing, and probing for vulnerabilities to gather information about target systems. The following classes from the dataset were included in this category: Back, Land, Neptune, Pod, Smurf, Teardrop, Apache2, Udp Storm, Processtable, and Worm.

**DoS (Denial of Service):** DoS attacks involve coordinated efforts to flood target systems with traffic, overwhelming them and causing service disruptions. The following classes from the dataset were included in this category: Satan, Ipsweep, Nmap, Portsweep, Mscan, and Saint.

**Unauthorized Access:** This class includes activities that aim to gain unauthorized access or control over systems, applications, or data. It encompasses various attacks and intrusion attempts, including user to root (U2R) and remote to local (R2L) attacks. The following classes from the dataset were included in this category: guess_passwd, ftp_write, Imap, Phf, Multihop, Warezmaster, Warezclient, Spy, Xlock, Xsnoop, Snmp Guess, Snmpgetattack, Httptunnel, Sendmail, Named, Mailbomb, buffer_overflow, Loadmodule, Rootkit, Perl, Sqlattack, Xterm, and Ps.

**Normal:** This class represents typical and benign network activities and was not associated with any attack types.

### 3.3 Data Split

The data was split into 67% and 33% training and testing datasets using stratification to ensure representation from each class. This method was chosen over random splitting to avoid biases and improve the model's accuracy on underrepresented groups. Stratification also helps to identify and address biases, reduce variance, and improve interpretability.

### 3.4 Model and Algorithm

The generic neural network (NN) algorithm was tested after the data processing techniques. One generic NN was kept as a control group without the dataset grouping shown in Table 1 to test the efficacy of the data processing techniques, i.e., a larger NN trained on all 23 classes to be compared with the proposed techniques and other machine learning and deep learning techniques. The models were then saved for future use, deployment, and reference.

The dataset was used to train four MLPs through different experiments. The first MLP was trained without clustering the labels. The second MLP was trained on the entire dataset with a weight attached to each class to establish a similar loss distribution among the different classes. The third MLP was trained on the proposed label categorization from Table 1 with a highly truncated model. Finally, the same truncated MLP was trained with newly curated weights for each class in Table 1. Since MLPs are advanced algorithms, the features in Table 2 were not dropped to ensure that the model gets the maximum number of features, which would help the model to make finer decision boundaries between two distinct classes.

The main advantage of the truncated models lies in their efficiency. The truncated models have 13892 parameters, whereas the base models have 62871. This size is achieved by decreasing neurons in hidden layers and removing certain hidden layers. Having a smaller model makes it more efficient and reduces the training speed.

### 3.5 Metrics

When evaluating a model, relying solely on accuracy can be flawed as it does not provide a complete picture of how well the model performs. To get a more accurate assessment, additional metrics beyond accuracy should be considered. Some of the important metrics that should be evaluated along with accuracy include precision, recall, F1 score, and confusion matrix. Precision measures a model's ability to identify relevant instances, while recall measures its ability to identify true positives. F1 score is a harmonic means of precision and recall, making it a good overall performance measure.

Confusion matrix is another valuable tool that provides insight into areas where the model makes mistakes. It shows how many instances from each class were predicted to belong to each class, giving an idea of the classes the model has biases for and against and the scale of the bias. This approach ensures that the model's performance is evaluated, considering each class's characteristics in the dataset.

### 3.6 SHAP Values

We trained a SHAP kernel explainer [12] on 50 randomly chosen data from the testing dataset. The explainer returns different SHAP values for each of the labels, determining the impact of each feature on the model's predictions. We used SHAP's built in tools to visualize the impact of each feature in the two truncated models.

## 4 Results

Neural networks were let to train for 20 epochs. After 20 epochs, each model was tested with 33 percent of testing dataset. For each epoch, the entire testing dataset was used as the validation dataset to validate the training process and intervene if the model was overfitting or underfitting.

As expected, and shown in Fig 2 and Fig 5, the training and validation loss for both the truncated models were the least, but the weighted loss for the truncated models were higher because of the better representation in the overall loss function.

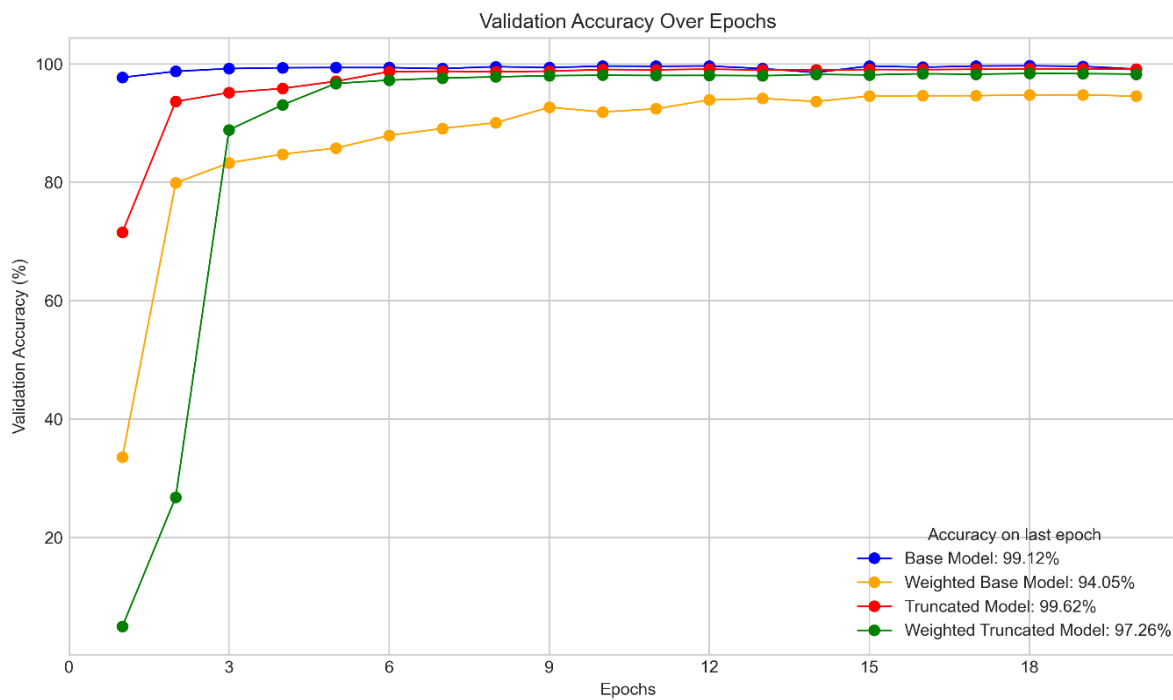

**Fig. 1.** Validation accuracy over 20 epochs of all models.

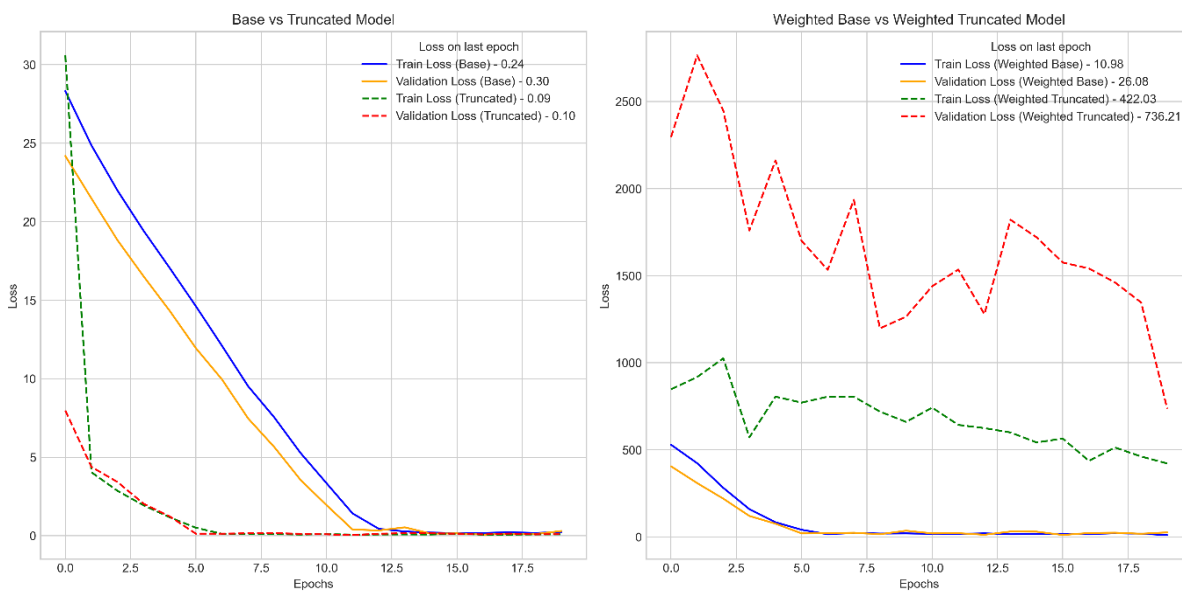

**Fig. 2.** Training and validation loss for weighted and unweighted models.

Notably, from Fig 1, Table II and III, the weighted truncated model outperformed the weighted base model, and the truncated model outperformed the base model regarding F1-score, recall, precision, accuracy, and loss.

Table 2. Classification report for base models with and without weights.

| Class | Base Model | | | Weighted Base Model | | | Support |
|---|---|---|---|---|---|---|---|
| | Precision | Recall | F1 Score | Precision | Recall | F1 score | |
| Normal | 0.9913 | 0.9728 | 0.9819 | 0.9969 | 0.7358 | 0.8467 | 321018 |
| Buffer overflow | 1 | 0 | 0 | 0 | 0 | 0 | 10 |
| Loadmodule | 1 | 0 | 0 | 1 | 0 | 0 | 3 |
| Perl | 1 | 0 | 0 | 0 | 0 | 0 | 1 |
| Neptune | 0.9961 | 0.9985 | 0.9973 | 0.9985 | 0.9775 | 0.9879 | 353766 |
| Smurf | 0.9998 | 0.9997 | 0.9998 | 0.9999 | 0.9992 | 0.9995 | 926603 |
| Guess passwd | 1 | 0 | 0 | 1 | 0 | 0 | 17 |
| Pod | 1 | 0 | 0 | 0.0073 | 0.3678 | 0.0142 | 87 |
| Teardrop | 1 | 0 | 0 | 0.0322 | 0.9009 | 0.0622 | 323 |
| Portsweep | 0.2501 | 0.8935 | 0.3909 | 0.1354 | 0.92 | 0.2361 | 3436 |
| Ipsweep | 0.8871 | 0.7208 | 0.7953 | 0.1817 | 0.9094 | 0.3029 | 4119 |
| Land | 1 | 0 | 0 | 1 | 0 | 0 | 7 |
| Ftp write | 1 | 0 | 0 | 0 | 0 | 0 | 3 |
| Back | 0.9855 | 0.1871 | 0.3145 | 0.0669 | 0.9849 | 0.1253 | 727 |
| Imap | 1 | 0 | 0 | 1 | 0 | 0 | 4 |
| Satan | 0.9605 | 0.8166 | 0.8827 | 0.9464 | 0.8249 | 0.8815 | 5244 |
| Phf | 1 | 0 | 0 | 1 | 0 | 0 | 1 |
| Nmap | 1 | 0 | 0 | 0.0341 | 0.0955 | 0.0502 | 764 |
| Multihop | 1 | 0 | 0 | 1 | 0 | 0 | 2 |
| Warezmaster | 0 | 0 | 0 | 0.0006 | 0.7143 | 0.0013 | 7 |
| Warezclient | 1 | 0 | 0 | 0.0053 | 0.3798 | 0.0105 | 337 |
| Spy | 1 | 0 | 0 | 1 | 0 | 0 | 1 |
| Rootkit | 1 | 0 | 0 | 0 | 0 | 0 | 3 |
| Accuracy | | 0.9912 | | | 0.9405 | | 1616483 |
| Macro average | 0.9161 | 0.243 | 0.2331 | 0.4959 | 0.383 | 0.1965 | 1616483 |
| Weighted average | 0.9953 | 0.9912 | 0.9922 | 0.9936 | 0.9405 | 0.9615 | 1616483 |

Table 3. Classification report of truncated and weighted truncated models.

| Class | Truncated Model | | | Truncated Weighted Model | | | Support |
|---|---|---|---|---|---|---|---|
| | Precision | Recall | F1 Score | Precision | Recall | F1 score | |
| Normal | 0.9908 | 0.996 | 0.9934 | 0.9958 | 0.9023 | 0.9468 | 321018 |
| Probe | 0.9986 | 0.9989 | 0.9987 | 0.9983 | 0.9907 | 0.9945 | 1281513 |
| DOS | 0.8842 | 0.7773 | 0.8273 | 0.3507 | 0.9482 | 0.512 | 13563 |
| Unauthorized | 1 | 0 | 0 | 0.0076 | 0.3368 | 0.0149 | 389 |
| Accuracy | | 0.9962 | | | 0.9726 | | 1616483 |
| Macro Average | 0.9684 | 0.693 | 0.7048 | 0.5881 | 0.7945 | 0.617 | 1616483 |
| Weighted Average | 0.9961 | 0.9962 | 0.996 | 0.9921 | 0.9726 | 0.9807 | 1616483 |

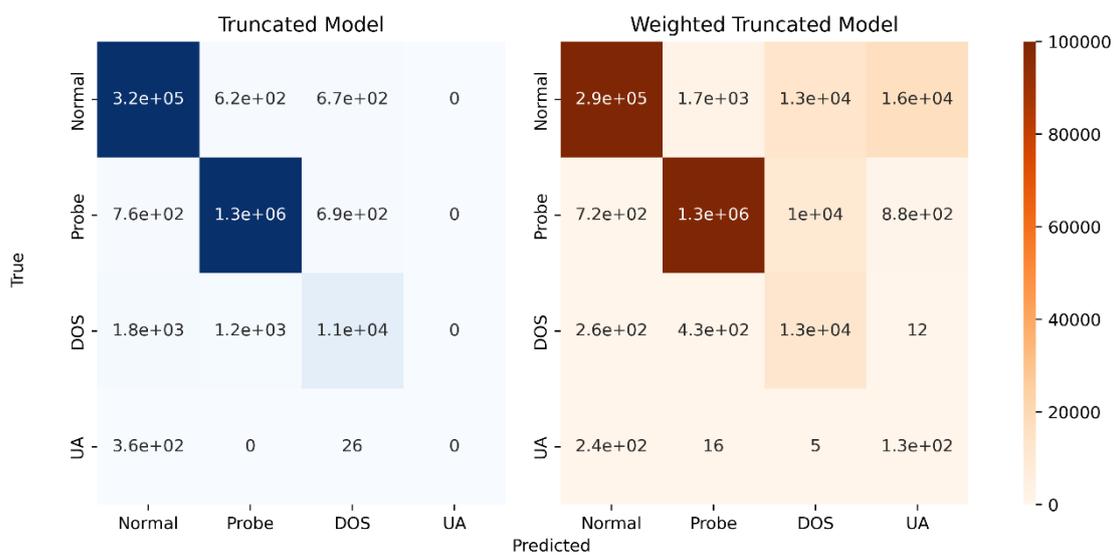

**Fig. 3.** Confusion matrix of truncated and weighted truncated models.

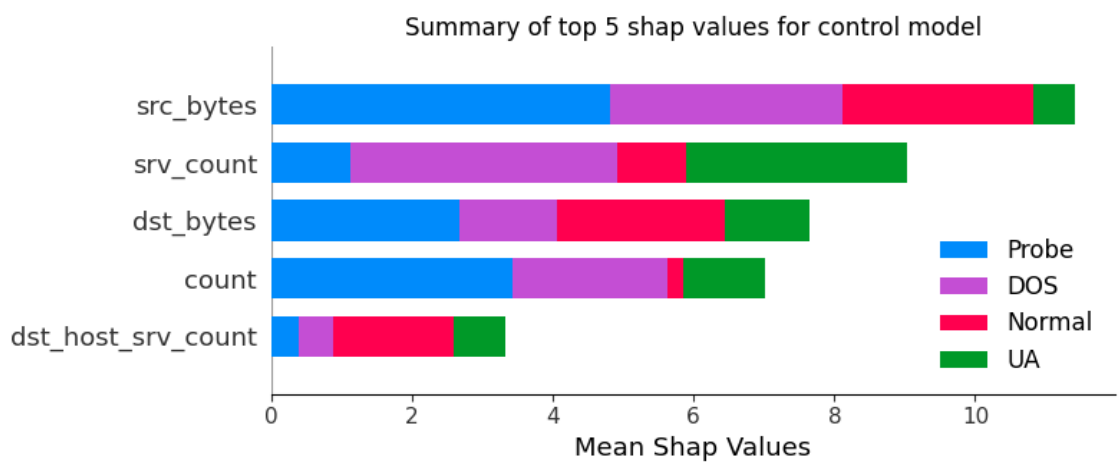

**Fig. 4.** Top 5 SHAP values for control model.

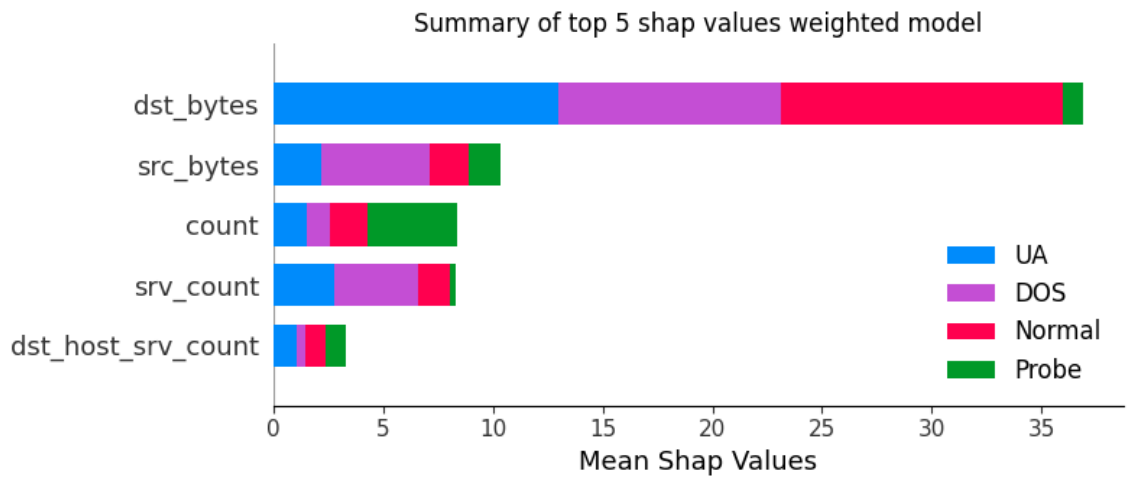

**Fig. 5.** Top 5 SHAP values for weighted model.

The base model performed well for certain classes with high numbers of support (samples) like Normal, Neptune, and Smurf with suitable precision, recall, and F1 scores. It scored acceptable metrics in classes Ipsweep and Satan. Except for Portsweep and Back, all other classes have an F1 score of 0. This means that either the precision or recall is 0 for these classes. The result is in line with our literature review on zero-day where models are accurate but there is low variance for classes with small number of samples. Although the weighted base model tries to overcome this fallacy by giving more emphasis to underrepresented classes during the training process, it was barely able to make any significant improvement. The maximum change in the unweighted control model, as shown in Table II is in F1-score of 0.01% in Warezclient while significantly lowering all other metrics in almost every other class. This indicates overfitting in both weighted and unweighted control models.

With a significantly lower number of classes, the smaller truncated models performed better than the base counterparts in terms of all metrics, reported in Table III, including precision, recall and F1 scores, for both the unweighted and the weighted models. From Fig. 1 and Table II and III, the truncated model improved the validation accuracy by 0.4% and the weighted counterpart increased it by 3.2%.

In the confusion matrix of truncated models in Fig 3, it can be noticed that the truncated models did not produce a single output in the target variable Unauthorized Access (UA) while the weighted model at least produced results in all of the classes and increased the recall from 0 to 0.33, depicted by the classification report in Table III.

The summary plot from the top 5 most impactful features, shown in Fig 5, gives an important insight into the thought process behind the model's reasoning. It can be observed that the top 5 features considered by both models in the KDD99 dataset are drastically different. Even in the graph, we can see most importance in SHAP values is given to the Probe class in the unweighted model, while the weighted model is dedicated to the most underrepresented Unauthorized Access class. The feature importance on output production depicted by Fig 5 further boosts the notion of how weights in a model compels it to think and observe features differently, thus producing a more robust and unbiased result. This solidifies our initial assumption that weighing the model can help to make it less biased towards underrepresented classes in IDS.

## 5    Conclusion and Future Considerations

The comparative analysis between models, such as the truncated and base weighted models, emphasized the significance of reorganizing or merging classes. The performance improvements observed in the truncated models, particularly in precision, recall, and F1 score metrics across diverse classes underscore the efficacy of such structural modifications in optimizing intrusion detection models for real-world applications. Adding weights in the models helps emphasize classes with a smaller number of samples, which hints towards a combined model approach using the truncated and weighted truncated models.

The efficiency of truncated models by reducing the number of trainable parameters shows promise in future research ideas. The two truncated and truncated weighted models have subtle differences that would have the potential for an improved system if combined. The SHAP values and use of XAI especially drive the concept behind an importance for ensemble learning method that could take advantage of the capabilities of both models and provide even more robust and dependable results in zero-day threat detection.

Future research could examine the potential of ensemble techniques like stacking with gradient boosting models or blending with adaptive weights based on contextual bandit algorithms. Investigating hierarchical models with specialized layers for different attack types could also be fruitful. Further investigation into combining these techniques is warranted to explore potential synergies and address trade-offs. Ultimately, these efforts can pave the way for more reliable and adaptable IDS that go beyond accuracy as the sole performance measure.

In conclusion, the study advocates for a refined and nuanced approach to IDS design, emphasizing the necessity of accounting for unique classes within datasets. By combining classes, exploring model variance, exploring the model's blackbox approach with XAI and proposing a potential combined approach, this research aims to pave the way for enhanced IDS reliability and performance in real-world scenarios that look beyond accuracy as the deterministic metric for judgment.

# 6 ACKNOWLEDGEMENT

This study was supported by the University of Southern Mississippi.